\documentclass{bmvc2k}
\usepackage{graphicx}
\usepackage{float}
\usepackage{url}
\usepackage{amsmath}
\usepackage{algorithm}
\usepackage{algpseudocode}
\usepackage{mathtools}
\usepackage{multirow}
\usepackage{makecell}

\title{Branched Multi-Task Networks: \\ Deciding What Layers To Share}

\addauthor{Simon Vandenhende}{simon.vandenhende@kuleuven.be}{1}
\addauthor{Stamatios Georgoulis}{stamatios.georgoulis@vision.ee.ethz.ch}{2}
\addauthor{Bert De Brabandere}{bert.debrabandere@esat.kuleuven.be}{1}
\addauthor{Luc Van Gool}{vangool@vision.ee.ethz.ch}{12}

\addinstitution{
 PSI-ESAT\\
 KU Leuven\\
 Leuven, Belgium
}
\addinstitution{
 CVL/TRACE\\
 ETH Zurich\\
 Zurich, Switzerland
}

\runninghead{Vandenhende et al.}{Branched Multi-Task Networks}

\def\etal{\emph{et al}\bmvaOneDot}

\begin{document}

\maketitle

\begin{abstract}
In the context of multi-task learning, neural networks with branched architectures have often been employed to jointly tackle the tasks at hand. Such ramified networks typically start with a number of shared layers, after which different tasks branch out into their own sequence of layers. Understandably, as the number of possible network configurations is combinatorially large, deciding what layers to share and where to branch out becomes cumbersome. Prior works have either relied on ad hoc methods to determine the level of layer sharing, which is suboptimal, or utilized neural architecture search techniques to establish the network design, which is considerably expensive. In this paper, we go beyond these limitations and propose an approach to automatically construct branched multi-task networks, by leveraging the employed tasks' affinities. Given a specific budget, i.e. number of learnable parameters, the proposed approach generates architectures, in which shallow layers are task-agnostic, whereas deeper ones gradually grow more task-specific. Extensive experimental analysis across numerous, diverse multi-tasking datasets shows that, for a given budget, our method consistently yields networks with the highest performance, while for a certain performance threshold it requires the least amount of learnable parameters.
\end{abstract}

\section{Introduction}
Deep neural networks are usually trained to tackle different tasks in isolation. Humans, in contrast, are remarkably good at solving a multitude of tasks concurrently. Biological data processing appears to follow a multi-tasking strategy too; instead of separating tasks and solving them in isolation, different processes seem to share the same early processing layers in the brain -- see e.g. V1 in macaques~\cite{gur2007direction}. Drawing inspiration from such observations, deep learning researchers began to develop multi-task networks with branched architectures.

As a whole, multi-task networks~\cite{caruana1997multitask} seek to improve generalization and processing efficiency through the joint learning of related tasks. Compared to the typical learning of separate deep neural networks for each of the individual tasks, multi-task networks come with several advantages. First, due to their inherent layer sharing~\cite{kokkinos2017ubernet,lu2017fully,kendall2018multi,guo2018dynamic,liu2019end}, the resulting memory footprint is typically substantially lower. Second, as features in the shared layers do not need to be calculated repeatedly for the different tasks, the overall inference speed is often higher~\cite{neven2017fast,lu2017fully}. Finally, multi-task networks may outperform their single-task counterparts~\cite{kendall2018multi,xu2018pad,sener2018multi,maninis2019attentive}. Evidently, there is merit in utilizing multi-task networks.

When it comes to designing them, however, a significant challenge is to decide on the layers that need to be shared among tasks. Assuming a hard parameter sharing setting\footnote{In this setting, the input is first encoded through a stack of shared layers, after which tasks branch out into their own sequence of task-specific layers~\cite{kokkinos2017ubernet,lu2017fully,kendall2018multi,guo2018dynamic,sener2018multi}. Alternatively, a set of task-specific networks can be used in conjunction with a feature sharing mechanism~\cite{misra2016cross,liu2019end,ruder2019latent}. The latter approach is termed soft parameter sharing in the literature.}, the number of possible network configurations grows quickly with the number of tasks. As a result, a trial-and-error procedure to define the optimal architecture becomes unwieldy. Resorting to neural architecture search~\cite{elsken2019neural} techniques is not a viable option too, as in this case, the layer sharing has to be jointly optimized with the layers types, their connectivity, etc., rendering the problem considerably expensive. Instead, researchers have recently explored more viable alternatives, like routing~\cite{rosenbaum2018routing}, stochastic filter grouping~\cite{bragman2019stochastic}, and feature partitioning~\cite{newell2019feature}, which are, however, closer to the soft parameter sharing setting. Previous works on hard parameter sharing opted for the simple strategy of sharing the initial layers in the network, after which all tasks branch out simultaneously. The point at which the branching occurs is usually determined ad hoc~\cite{kendall2018multi,guo2018dynamic,sener2018multi}. This situation hurts performance, as a suboptimal grouping of tasks can lead to the sharing of information between unrelated tasks, known as \emph{negative transfer}~\cite{zhao2018modulation}.

In this paper, we go beyond the aforementioned limitations and propose a novel approach to decide on the degree of layer sharing between multiple visual recognition tasks in order to eliminate the need for manual exploration. To this end, we base the layer sharing on measurable levels of \emph{task affinity} or \emph{task relatedness}: two tasks are strongly related, if their single task models rely on a similar set of features. \cite{zamir2018taskonomy} quantified this property by measuring the performance when solving a task using a variable sets of layers from a model pretrained on a different task. However, their approach is considerably expensive, as it scales quadratically with the number of tasks. Recently,~\cite{dwivedi2019representation} proposed a more efficient alternative that uses representation similarity analysis (RSA) to obtain a measure of task affinity, by computing correlations between models pretrained on different tasks. Given a dataset and a number of tasks, our approach uses RSA to assess the task affinity at arbitrary locations in a neural network. The task affinity scores are then used to construct a branched multi-task network in a fully automated manner. In particular, our task clustering algorithm groups similar tasks together in common branches, and separates dissimilar tasks by assigning them to different branches, thereby reducing the negative transfer between tasks. Additionally, our method allows to trade network complexity against task similarity. We provide extensive empirical evaluation of our method, showing its superiority in terms of multi-task performance vs computational resources.  

\section{Related work}

\textbf{Multi-task learning.}
Multi-task learning (MTL)~\cite{caruana1997multitask,ruder2017overview} is associated with the concept of jointly learning multiple tasks under a single model. This comes with several advantages, as described above. Early work on MTL often relied on sparsity constraints~\cite{yuan2006model,argyriou2007multi,lounici2009taking,jalali2010dirty,liu2017distributed} to select a small subset of features that could be shared among all tasks. However, this can lead to negative transfer when not all tasks are related to each other. A general solution to this problem is to cluster tasks based on prior knowledge about their similarity or relatedness~\cite{evgeniou2004regularized,abernethy2009new,agarwal2010learning,zhou2011malsar,kumar2012learning}. 

In the deep learning era, MTL models can typically be classified as utilizing soft or hard parameter sharing. In soft parameter sharing, each task is assigned its own set of parameters and a feature sharing mechanism handles the cross-task talk. Cross-stitch networks~\cite{misra2016cross} softly share their features among tasks, by using a linear combination of the activations found in multiple single task networks. Sluice networks~\cite{ruder2019latent} extend cross-stitch networks and allow to learn the selective sharing of layers, subspaces and skip connections. In a different vein, multi-task attention networks~\cite{liu2019end} use an attention mechanism to share a general feature pool amongst task-specific networks. In general, MTL networks using soft parameter sharing are limited in terms of scalability, as the size of the network tends to grow linearly with the number of tasks.

In hard parameter sharing, the parameter set is divided into shared and task-specific parameters. MTL models using hard parameter sharing are often based on a generic framework with a shared off-the-shelf encoder, followed by task-specific decoder networks~\cite{neven2017fast,kendall2018multi,chen2018gradnorm,sener2018multi}. Multilinear relationship networks~\cite{long2017learning} extend this framework by placing tensor normal priors on the parameter set of the fully connected layers. \cite{guo2018dynamic} proposed the construction of a hierarchical network, which predicts increasingly difficult tasks at deeper layers. A limitation of the aforementioned approaches is that the branching points are determined ad hoc, which can easily lead to negative transfer if the predefined task groupings are suboptimal. In contrast, in our branched multi-task networks, the degree of layer sharing is automatically determined in a principled way, based on task affinities. 

Our work bears some similarity to fully-adaptive feature sharing~\cite{lu2017fully} (FAFS), which starts from a thin network where tasks initially share all layers, but the final one, and dynamically grows the model in a greedy layer-by-layer fashion. Task groupings, in this case, are decided on the probability of concurrently simple or difficult examples across tasks. Differently, (1) our method clusters tasks based on feature affinity scores, rather than example difficulty, which is shown to achieve better results for a variety of datasets; (2) the tree structure is determined offline using the precalculated affinities for the whole network, and not online in a greedy layer-by-layer fashion, which promotes task groupings that are optimal in a global, rather than local, sense.

\textbf{Neural architecture search.}
Neural architecture search (NAS)~\cite{elsken2019neural} aims to automate the construction of the network architecture. Different algorithms can be characterized based on their search space, search strategy or performance estimation strategy. Most existing works on NAS, however, are limited to task-specific models~\cite{zoph2017neural,liu2018hierarchical,pham2018efficient,liu2018progressive,real2019regularized}. This is to be expected as when using NAS for MTL, layer sharing has to be jointly optimized with the layers types, their connectivity, etc., rendering the problem considerably expensive. To alleviate the heavy computation burden, a recent work~\cite{liang2018evolutionary} implemented an evolutionary architecture search for multi-task networks, while other researchers explored more viable alternatives, like routing~\cite{rosenbaum2018routing}, stochastic filter grouping~\cite{bragman2019stochastic}, and feature partitioning~\cite{newell2019feature}. In contrast to traditional NAS, the proposed methods do not build the architecture from scratch, but rather start from a predefined backbone network for which a layer sharing scheme is automatically determined.

\textbf{Transfer learning.}
Transfer learning~\cite{pan2010survey} uses the knowledge obtained when solving one task, and applies it to a different but related task. Our work is loosely related to transfer learning, as we use it to measure levels of task affinity. \cite{zamir2018taskonomy} provided a taxonomy for task transfer learning to quantify such relationships. However, their approach scales unfavorably w.r.t. the number of tasks, and we opted for a more efficient alternative proposed by~\cite{dwivedi2019representation}. The latter uses RSA to obtain a measure of task affinity, by computing correlations between models pretrained on different tasks. In our method, we use the performance metric from their work to compare the usefulness of different feature sets for solving a particular task. 

\textbf{Loss weighting.}
One of the known challenges of jointly learning multiple tasks is properly weighting the loss functions associated with the individual tasks. Early work~\cite{kendall2018multi} used the homoscedastic uncertainty of each task to weigh the losses. Gradient normalization~\cite{chen2018gradnorm} balances the learning of tasks by dynamically adapting the gradient magnitudes in the network. Liu~\etal~\cite{liu2019end} weigh the losses to match the pace at which different tasks are learned. Dynamic task prioritization~\cite{guo2018dynamic} prioritizes the learning of difficult tasks. \cite{sener2018multi} cast MTL as a multi-objective optimization problem, with the overall objective of finding a Pareto optimal solution. Note that, addressing the loss weighting issue in MTL is out of the scope of this work. In fact, all our experiments are based on a simple uniform loss weighing scheme.


\section{Method}
\label{sec: method}

\begin{figure}[t]
\begin{center}
\begin{minipage}[t]{.70\textwidth}
  \begin{center}
  \includegraphics[width=\textwidth]{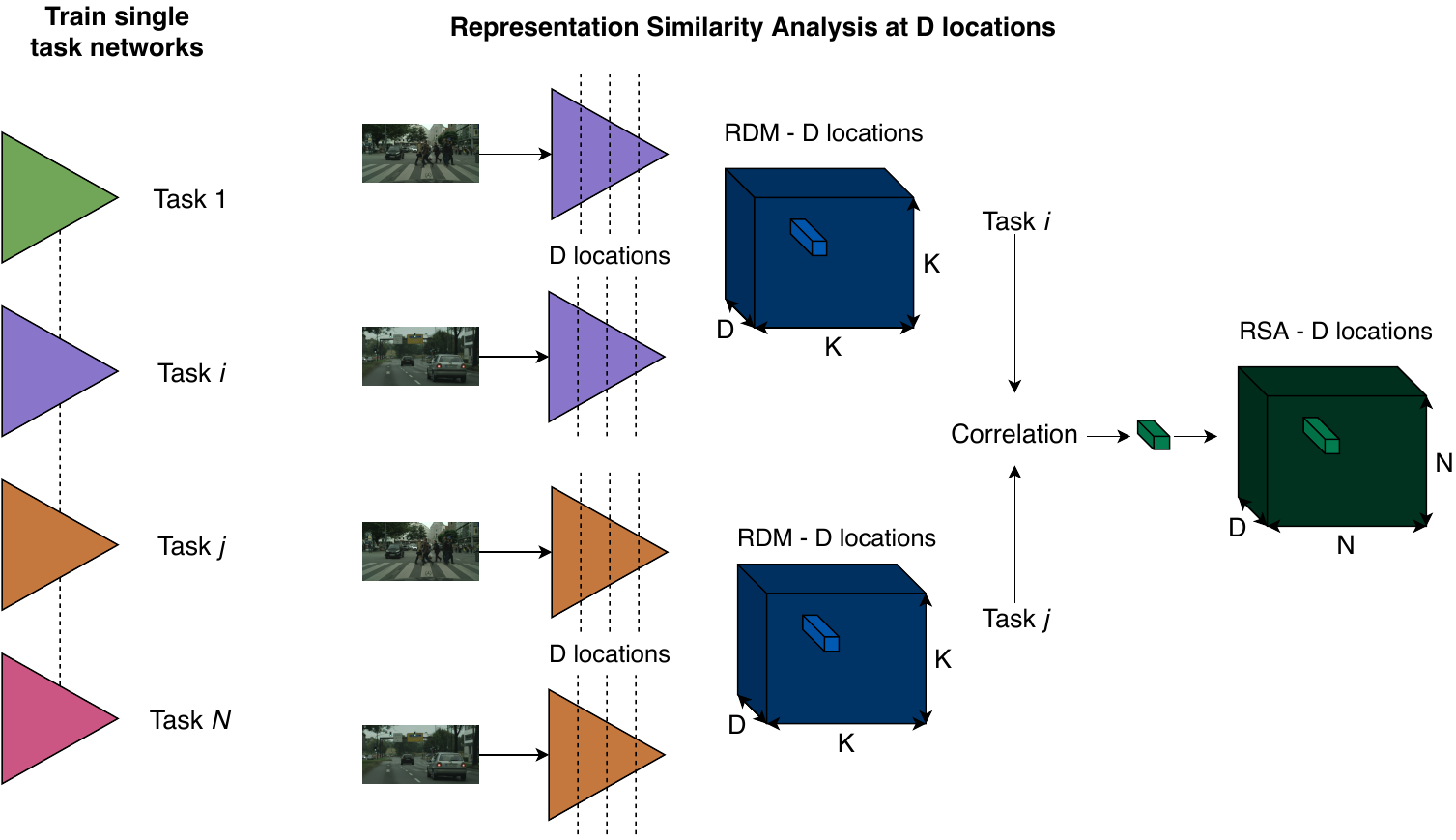}
  \end{center}
  \caption{Pipeline overview. (left) We train a single-task model for every task $t \in \mathcal{T}$. (middle) We use RSA to measure the task affinity at $D$ predefined locations in the sharable encoder. In particular, we calculate the representation dissimilarity matrices (RDM) for the features at $D$ locations using $K$ images, which gives a $D \times K \times K$ tensor per task. (right) The affinity tensor $\mathbf{A}$ is found by calculating the correlation between the RDM matrices, which results in a three-dimensional tensor of size $D \times N \times N$, with $N$ the number of tasks.}
  \label{fig:rsa}
\end{minipage}%
\hspace*{\fill}
\begin{minipage}[t]{.25\textwidth}
  \begin{center}
  \includegraphics[width=\textwidth]{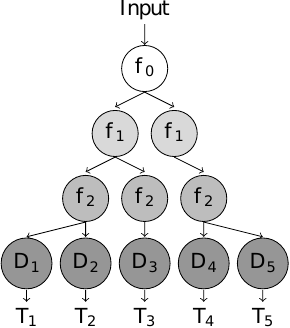}
  \end{center}
  \caption{Our pipeline's output is a branched multi-task network, similar to how NAS techniques output sample architectures. An example branched multi-task network is visualized here.}
  \label{fig:tree}
\end{minipage}
\end{center}
\caption{The proposed method: (a) calculate task affinities at various locations in the sharable encoder; (b) build a branched multi-task network based on the computed affinities.}
\label{fig: method}
\end{figure}

In this paper, we aim to jointly solve $N$ different visual recognition tasks $\mathcal{T}=\left\{t_1, \ldots, t_N\right\}$ given a computational budget $\mathcal{C}$, i.e. number of parameters or FLOPS. Consider a backbone architecture: an encoder, consisting of a sequence of shared layers or blocks $f_l$, followed by a decoder with a few task-specific layers. We assume an appropriate structure for layer sharing to take the shape of a tree. In particular, the first layers are shared by all tasks, while later layers gradually split off as they show more task-specific behavior. The proposed method aims to find an effective task grouping for the sharable layers $f_l$ of the encoder, i.e. grouping related tasks together in the same branches of the tree. When two tasks are strongly related, we expect their single-task models to rely on a similar feature set~\cite{zamir2018taskonomy}. Based on this viewpoint, the proposed method derives a task affinity score at various locations in the sharable encoder. The number of locations $D$ can be freely determined as the number of candidate branching locations. As such, the resulting task affinity scores are used for the automated construction of a branched multi-task network that fits the computational budget $\mathcal{C}$. Fig.~\ref{fig: method} illustrates our pipeline, while Algorithm~\ref{alg: algorithm} summarizes the whole procedure. 

\begin{algorithm}[t]
\caption{Branched Multi-Task Networks - Task clustering}
\label{alg: algorithm}
\small{
\begin{algorithmic}[1]
\State \textbf{Input:} Tasks $\mathcal{T}$, $K$ images $\mathcal{I}$, a sharable encoder $E$ with $D$ locations where we can branch, a set of task specific decoders $D_t$ and a computational budget $\mathcal{C}$.
\For{$t$ in $\mathcal{T}$} 
\State Train the encoder $E$ and task-specific decoder $D_t$ for task $t$.
\State $RDM^{t} \leftarrow RDM(E, D, \mathcal{I})$ \Comment{RDM for task $t$.}
\EndFor
\State $A_{d,i,j} \leftarrow r_s\left( \text{triu} \left( RDM^{t_i}_{d,:,:} \right), \text{triu} \left( RDM^{t_j}_{d,:,:}  \right) \right) $ \textbf{for} $t_i, t_j$ in $\mathcal{T}$ and $d$ in locations \Comment{Task affinity}
\State $D = 1 - A$ \Comment{Task dissimilarity}
\State \textbf{Return:} Task-grouping with minimal task dissimilarity that fits within $\mathcal{C}$
\end{algorithmic}
}
\end{algorithm}

\subsection{Calculate task affinity scores}

As mentioned, we rely on RSA to measure task affinity scores. This technique has been widely adopted in the field of neuroscience to draw comparisons between behavioral models and brain activity. Inspired by how~\cite{dwivedi2019representation} applied RSA to select tasks for transfer learning, we use the technique to assess the task affinity at predefined locations in the sharable encoder. Consequently, using the measured levels of task affinity, tasks are assigned in the same or different branches of a branched multi-task network, subject to the computational budget $\mathcal{C}$. 

The procedure to calculate the task affinity scores is the following. As a first step, we train a single-task model for each task $t_{i} \in \mathcal{T}$. The single-task models use an identical encoder $E$ - made of all sharable layers $f_l$ - followed by a task-specific decoder $D_{t_{i}}$. The decoder contains only task-specific operations and is assumed to be significantly smaller in size compared to the encoder. As an example, consider jointly solving a classification and a dense prediction task. Some fully connected layers followed by a softmax operation are typically needed for the classification task, while an additional decoding step with some upscaling operations is required for the dense prediction task. Of course, the appropriate loss functions are applied in each case. Such operations are part of the task-specific decoder $D_{t_{i}}$. The different single-task networks are trained under the same conditions.

At the second step, we choose $D$ locations in the sharable encoder $E$ where we calculate a two-dimensional task affinity matrix of size $N \times N$. When concatenated, this results in a three-dimensional tensor $\mathbf{A}$ of size $D \times N \times N$ that holds the task affinities at the selected locations. To calculate these task affinities, we have to compare the representation dissimilarity matrices (RDM) of the single-task networks - trained in the previous step - at the specified $D$ locations. To do this, a held-out subset of $K$ images is required. The latter images serve to compare the dissimilarity of their feature representations in the single-task networks for every pair of images. Specifically, for every task $t_i$, we characterize these learned feature representations at the selected locations by filling a tensor of size $D \times K \times K$. This tensor contains the dissimilarity scores $1-\rho$ between feature representations, with $\rho$ the Pearson correlation coefficient. Specifically, $\mathbf{RDM}_{d,i,j}$ is found by calculating the dissimilarity score between the features at location $d$ for image $i$ and $j$. The 3-D tensors are linearized to 1-D tensors to calculate the pearson correlation coefficient.

For a specific location $d$ in the network, the computed RDMs are symmetrical, with a diagonal of zeros. For every such location, we measure the similarity between the upper or lower triangular part of the RDMs belonging to the different single-task networks. We use the Spearman's correlation coefficient $r_s$ to measure similarity. When repeated for every pair of tasks, at a specific location $d$, the result is a symmetrical matrix of size $N \times N$, with a diagonal of ones. Concatenating over the $D$ locations in the sharable encoder, we end up with the desired task affinity tensor of size $D \times N \times N$. Note that, in contrast to prior work~\cite{lu2017fully}, the described method focuses on the features used to solve the single tasks, rather than the examples and how easy or hard they are across tasks, which is shown to result in better task groupings in Section~\ref{sec: experiments}. Furthermore, the computational overhead to determine the task affinity scores based on feature correlations is negligible. We conclude that the computational cost of the method boils down to pre-training $N$ single task networks. A detailed computational cost analysis can be found in the supplementary materials.

Other measures of task similarity~\cite{achille2019task2vec,cui2018large} probed the features from a network pre-trained on ImageNet. This avoids the need to pre-train a set of single-task networks first. However, in this case, the task dictionaries only consisted of various, related classification problems. Differently, we consider more diverse, and loosely related task (see Section~\ref{sec: experiments}). In our case, it is arguably more important to learn the task-specific information needed to solve a task. This motivates the use of pre-trained single-task networks. 

\subsection{Construct a branched multi-task network}
\label{sec: branching}

Given a computational budget $\mathcal{C}$, we need to derive how the layers (or blocks) $f_l$ in the sharable encoder $E$ should be shared among the tasks in $\mathcal{T}$. Each layer $f_l \in E$ is represented as a node in the tree, i.e. the root node contains the first layer $f_0$, and nodes at depth $l$ contain layer(s) $f_l$. The granularity of the layers $f_l$ corresponds to the intervals at which we measure the task affinity in the sharable encoder, i.e. the $D$ locations. When the encoder is split into $b_{l}$ branches at depth $l$, this is equivalent to a node at depth $l$ having $b_{l}$ children. The leaves of the tree contain the task-specific decoders $D_t$. Fig.~\ref{fig:tree} shows an example of such a tree using the aforementioned notation. Each node is responsible for solving a unique subset of tasks.

The branched multi-task network is built with the intention to separate dissimilar tasks by assigning them to separate branches. To this end, we define the dissimilarity score between two tasks $t_i$ and $t_j$ at location $d$ as $1-\mathbf{A}_{d,i,j}$, with $\mathbf{A}$ the task affinity tensor\footnote{This is not to be confused with the dissimilarity score used to calculate the RDM elements $\mathbf{RDM}_{d,i,j}$.}. The branched multi-task network is found by minimizing the sum of the task dissimilarity scores at every location in the sharable encoder. In contrast to prior work~\cite{lu2017fully}, the task affinity (and dissimilarity) scores are calculated a priori. This allows us to determine the task clustering offline. Since the number of tasks is finite, we can enumerate all possible trees that fall within the given computational budget $\mathcal{C}$. Finally, we select the tree that minimizes the task dissimilarity score. The task dissimilarity score of a tree is defined as $C_{cluster} = \sum_{l} C_{cluster}^{l}$, where $C_{cluster}^{l}$ is found by averaging the maximum distance between the dissimilarity scores of the elements in every cluster. The use of the maximum distance encourages the separation of dissimilar tasks. By taking into account the clustering cost at all depths, the procedure can find a task grouping that is considered optimal in a global sense. This is in contrast to the greedy approach in~\cite{lu2017fully}, which only minimizes the task dissimilarity locally, i.e. at isolated locations in the network. 



\section{Experiments}
\label{sec: experiments}
In this section, we evaluate the proposed method on a number of diverse multi-tasking datasets, that range from real to semi-real data, from few to many tasks, from dense prediction to classification tasks, and so on. For every experiment, we describe the most important elements of the setup. We report the number of parameters (\#P) for every model to facilitate a fair comparison. Additional implementation details can be found in the supplementary materials.

\subsection{Cityscapes}

\textbf{Dataset.}
The Cityscapes dataset~\cite{cordts2016cityscapes} considers the scenario of urban scene understanding. The train, validation and test set contain respectively 2975, 500 and 1525 real images, taken by driving a car in Central European cities. It considers a few dense prediction tasks: semantic segmentation (S), instance segmentation (I) and monocular depth estimation (D). As in prior works~\cite{kendall2018multi, sener2018multi}, we use a ResNet-50 encoder with dilated convolutions, followed by a Pyramid Spatial Pooling (PSP)~\cite{he2015spatial} decoder. Every input image is rescaled to 512 x 256 pixels. We reuse the approach from~\cite{kendall2018multi} for the instance segmentation task, i.e. we consider the proxy task of regressing each pixel to the center of the instance it belongs to. 

\begin{figure}[t]
\begin{minipage}[t]{.48\linewidth}
   \begin{center}
   \includegraphics[width=\textwidth]{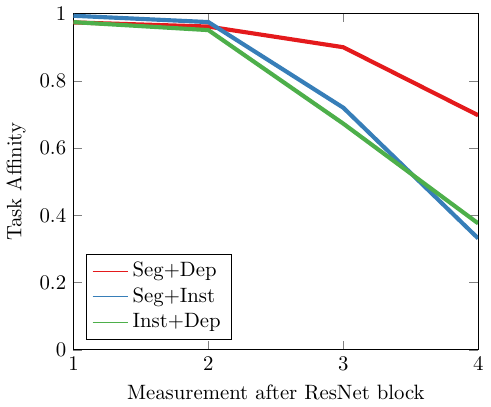}
   \end{center}
   \caption{Task affinity scores measured after each ResNet-50 block on Cityscapes.}
   \label{fig: cityscapes_affinity}
 \end{minipage}%
 \hspace*{\fill}
 \begin{minipage}[t]{.48\linewidth}
   \begin{center}
   \includegraphics[width=\textwidth]{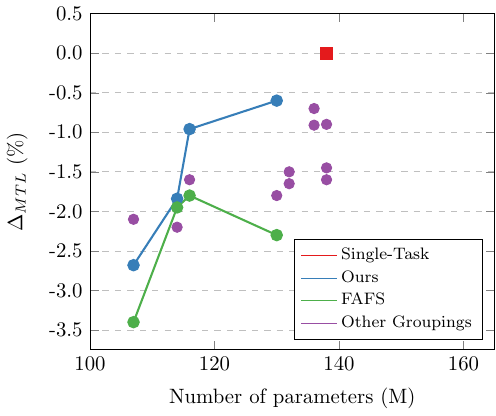}
   \end{center}
   \caption{Number of parameters versus multi-task performance on Cityscapes for different task groupings. The 'Other Groupings' contain any remaining tree structures that can be found by randomly branching the model in the last three ResNet blocks.}
   \label{fig: cityscapes}
 \end{minipage}
 \end{figure}

\noindent\textbf{Results.}
We measure the task affinity after every block (1 to 4) in the ResNet-50 model (see Fig.~\ref{fig: cityscapes_affinity}). The task affinity decreases in the deeper layers, due to the features becoming more task-specific. We compare the performance of the task groupings generated by our method with those by other approaches. As in~\cite{maninis2019attentive}, the performance of a multi-task model $m$ is defined as the average per-task performance drop/increase w.r.t. a single-task baseline $b$.

We trained all possible task groupings that can be derived from branching the model in the last three ResNet blocks. Fig.~\ref{fig: cityscapes} visualizes performance vs number of parameters for the trained architectures. Depending on the available computational budget $\mathcal{C}$, our method generates a specific task grouping. We visualize these generated groupings as a blue path in Fig.~\ref{fig: cityscapes}, when gradually increasing the computational budget $\mathcal{C}$. Similarly, we consider the task groupings when branching the model based on the task affinity measure proposed by FAFS~\cite{lu2017fully} (green path). We find that, in comparison, the task groupings devised by our method achieve higher performance within a given computational budget $\mathcal{C}$. Furthermore, in the majority of cases, for a fixed budget $\mathcal{C}$ the proposed method is capable of selecting the best performing task grouping w.r.t. performance vs parameters metric (blue vs other). 

We also compare our branched multi-task networks with cross-stitch networks~\cite{misra2016cross}, NDDR-CNNs~\cite{gao2019nddr} and MTAN~\cite{liu2019end} in Table~\ref{tab: cityscapes}. While cross-stitch nets and NDDR-CNNs give higher multi-task performance, attributed to their computationally expensive soft parameter sharing setting, our branched networks can strike a better trade-off between the performance and number of parameters. In particular, we can effectively sample architectures which lie between the extremes of a baseline multi-task model and a cross-stitch or NDDR-CNN architecture. Finally, our models provide a more computationally efficient alternative to the MTAN model, which reports similar performance while using more parameters.

\subsection{Taskonomy}

\textbf{Dataset.}
The Taskonomy dataset~\cite{zamir2018taskonomy} contains semi-real images of indoor scenes, annotated for 26 (dense prediction, classification, etc.) tasks. Out of the available tasks, we select scene categorization (C), semantic segmentation (S), edge detection (E), monocular depth estimation (D) and keypoint detection (K). The task dictionary was selected to be as diverse as possible, while still keeping the total number of tasks reasonable for all computations. We use the tiny split of the dataset, containing 275k train, 52k validation and 54k test images. We reuse the architecture and training setup from~\cite{zamir2018taskonomy}: the encoder is based on ResNet-50; a 15-layer fully-convolutional decoder is used for the pixel-to-pixel prediction tasks.

\begin{table}[t]
\begin{center}
\scriptsize{
\begin{tabular}{|l|c|c|c|c|c|c|c|}
\hline
\hline
Method & \multirowcell{2}[0pt][c]{D\\(L1)$\downarrow$} &
\multirowcell{2}[0pt][c]{S\\(IoU)$\uparrow$} & 
\multirowcell{2}[0pt][c]{C\\(Top-5)$\uparrow$} &
\multirowcell{2}[0pt][c]{E\\(L1)$\downarrow$} &
\multirowcell{2}[0pt][c]{K\\(L1)$\downarrow$} &
\multirowcell{2}[0pt][c]{\#P\\(M)} &
\multirowcell{2}[0pt][c]{$\Delta_{MTL}$\\(\%)$\uparrow$} \\
& & & & & & & \\
\hline
Single task & 0.60 & 43.5 & 66.0 & 0.99 & 0.23 & 224 & + 0.00 \\
MTL baseline & 0.75 & 47.8 & 56.0 & 1.37 & 0.34 & 130 & - 22.50 \\
MTAN & 0.71 & 43.8 & 59.6 & 1.86 & 0.40 & 158 & -37.36 \\
Cross-stitch & 0.61 & 44.0 & 58.2 & 1.35 & 0.50 & 224 & - 32.29 \\
NDDR-CNN & 0.66 & 45.9 & 64.5 & 1.05 & 0.45 & 258 & - 21.02 \\
\hline
FAFS - 1 & 0.74 & 46.1 & 62.7 & 1.30 & 0.39 & 174 & - 24.5 \\
FAFS - 2 & 0.80 & 39.9 & 62.4 & 1.68 & 0.52 & 188 & - 48.32 \\
FAFS - 3 & 0.74 & 46.1 & 64.9 & 1.05 & 0.27 & 196 & - 8.48 \\
\hline
Ours - 1 & 0.76 & 47.6 & 63.3 & 1.12 & 0.29 & 174 & - 11.88 \\
Ours - 2 & 0.74 & 48.0 & 63.6 & 0.96 & 0.35 & 188 & - 12.66 \\
Ours - 3 & 0.74 & 47.9 & 64.5 & 0.94 & 0.26 & 196 & - 4.93 \\
\hline
\end{tabular}
}
\end{center}
\caption{Results on the tiny Taskonomy test set. The results for edge (E) and keypoints (K) detection were multiplied by a factor of 100 for better readability. The FAFS models refer to generating the task groupings with the task affinity technique proposed by~\cite{lu2017fully}.}
\label{tab: taskonomy}
\end{table}

\noindent\textbf{Results.}
The task affinity is again measured after every ResNet block. Since the number of tasks increased to five, it is very expensive to train all task groupings exhaustively, as done above. Instead, we limit ourselves to three architectures that are generated when gradually increasing the parameter budget. As before, we compare our task groupings against the method from~\cite{lu2017fully}. The numerical results can be found in Table~\ref{tab: taskonomy}. The task groupings themselves are shown in the supplementary materials.

The effect of the employed task grouping technique can be seen from comparing the performance of our models against the corresponding FAFS models, generated by~\cite{lu2017fully}. The latter are consistently outperformed by our models. Compared to the results on Cityscapes (Fig.~\ref{fig: cityscapes}), we find that the multi-task performance is much more susceptible to the employed task groupings, possibly due to negative transfer. Furthermore, we observe that none of the soft parmeter sharing models can handle the larger, more diverse task dictionary: the performance decreases when using these models, while the number of parameters increases. This is in contrast to our branched multi-task networks, which seem to handle the diverse set of tasks rather positively. As opposed to~\cite{zamir2018taskonomy}, but in accordance with~\cite{maninis2019attentive}, we show that it is possible to solve many heterogeneous tasks simultaneously when the negative transfer is limited, by separating dissimilar tasks from each other in our case. In fact, our approach is the first to show such consistent performance across different multi-tasking scenarios and datasets. Existing approaches seem to be tailored for particular cases, e.g. few/correlated tasks, synthetic-like data, binary classification only tasks, etc., whereas we show stable performance across the board of different experimental setups.

\subsection{CelebA}

\textbf{Dataset.}
The CelebA dataset~\cite{liu2015faceattributes} contains over 200k images of celebrities, labeled with 40 facial attribute categories. The training, validation and test set contain 160k, 20k and 20k images respectively. We treat the prediction of each facial attribute as a single binary classification task, as in~\cite{lu2017fully,sener2018multi,huang2018gnas}. To ensure a fair comparison: we reuse the thin-$\omega$ model from~\cite{lu2017fully} in our experiments; the parameter budget $\mathcal{C}$ is set for the model to have the same amount of parameters as prior work. 

\begin{table}[t]
\begin{minipage}[t]{.52\linewidth}
 \begin{center}
 \scriptsize{
     \begin{tabular}{|l|c|c|c|c|c|}
     \hline
     Method & \multirowcell{2}[0pt][c]{S\\(IoU)$\uparrow$} & \multirowcell{2}[0pt][c]{I\\(px)$\downarrow$} &
     \multirowcell{2}[0pt][c]{D\\(px)$\downarrow$} & \multirowcell{2}[0pt][c]{\#P\\(M)} & \multirowcell{2}[0pt][c]{$\Delta_{MTL}$\\(\%)$\uparrow$} \\
      & & & & & \\
     \hline
Single task & 65.2 & 11.7 & 2.57 & 138 & +0.00 \\
MTL baseline & 61.5 & 11.8 & 2.66 & 92 & -3.33 \\
MTAN & 62.8 & 11.8 & 2.66 & 113 & -2.53 \\
Cross-stitch & 65.1 & 11.6 & 2.55 & 140 & +0.42 \\
NDDR-CNN & 65.6 & 11.6 & 2.54 & 190 & +0.89 \\
    \hline
Ours - 1 & 62.1 & 11.7 & 2.66 & 107 & -2.68 \\
Ours - 2 & 62.7 & 11.7 & 2.62 & 114 & -1.84 \\
Ours - 3 & 64.1 & 11.6 & 2.62 & 116 & -0.96 \\
     \hline
     \end{tabular}}
 \end{center}
 \caption{Results on the Cityscapes validation set.}
  \label{tab: cityscapes}
\end{minipage}
\hspace*{\fill}
\begin{minipage}[t]{.47\linewidth}
\begin{center}
\scriptsize{
\begin{tabular}{|l|c|c|}
\hline
Method & Acc. (\%) & \#P (M) \\
\hline
MOON \cite{rudd2016moon} & 90.94 & 119.73 \\
Independent Group \cite{hand2017attributes} & 91.06 & - \\ 
MCNN \cite{hand2017attributes} & 91.26 & - \\
MCNN-AUX \cite{hand2017attributes} & 91.29 & - \\
VGG-16 \cite{lu2017fully} & 91.44 & 134.41 \\
\textbf{FAFS} \cite{lu2017fully} &  \textbf{90.79} & \textbf{2.09} \\
\textit{GNAS} \cite{hand2017attributes} & \textit{91.63} & \textit{7.73} \\
Res-18 (Uniform) \cite{sener2018multi} & 90.38 & 11.2 \\
Res-18 (MGDA) \cite{sener2018multi} & 91.75 & 11.2 \\
\hline
\textbf{Ours-32}& \textbf{91.46} & \textbf{2.20} \\
\textit{Ours-64} & \textit{91.73} & \textit{7.73} \\
\hline
\end{tabular}
}
\end{center}
\caption{Results on the CelebA test set. The \textbf{Ours-32}, \textit{Ours-64} architectures are found by optimizing the task clustering for the parameter budget that is used in the \textbf{FAFS}, \textit{GNAS} model respectively.}
\label{tab:celeba}
\end{minipage}
\end{table}

\noindent\textbf{Results.}
Table~\ref{tab:celeba} shows the results on the CelebA test set. Our branched multi-task networks outperform earlier works~\cite{lu2017fully, huang2018gnas} when using a similar amount of parameters. Since the Ours-32 model (i.e. $\omega$ is 32) only differs from the FAFS model on the employed task grouping technique, we can conclude that the proposed method devises more effective task groupings for the attribute classification tasks on CelebA. Furthermore, the Ours-32 model performs on par with the VGG-16 model, while using 64 times less parameters. We also compare our results with the ResNet-18 model from~\cite{sener2018multi}. The Ours-64 model performs 1.35\% better compared to the ResNet-18 model when trained with a uniform loss weighing scheme. More noticeably, the Ours-64 model performs on par with the state-of-the-art ResNet-18 model that was trained with the MGDA loss weighing scheme from~\cite{sener2018multi}, while at the same time using 31\% less parameters (11.2 vs 7.7 M). 


\section{Conclusion}
In this paper, we introduced a principled approach to automatically construct branched multi-task networks for a given computational budget. To this end, we leverage the employed tasks' affinities as a quantifiable measure for layer sharing. The proposed approach can be seen as an abstraction of NAS for MTL, where only layer sharing is optimized, without having to jointly optimize the layers types, their connectivity, etc., as done in traditional NAS, which would render the problem considerably expensive. Extensive experimental analysis shows that our method outperforms existing ones w.r.t. the important metric of multi-tasking performance vs number of parameters, while at the same time showing consistent results across a diverse set of multi-tasking scenarios and datasets.  

\noindent\textbf{Acknowledgment} This work is sponsored by the Flemish Government under the Artificiele Intelligentie (AI) Vlaanderen programme. The authors also acknowledge support by Toyota via the TRACE project and MACCHINA (KU Leuven, C14/18/065).

\clearpage

\setcounter{section}{0}
\renewcommand\thesection{\Alph{section}}
\setcounter{figure}{0}
\setcounter{table}{0}
\renewcommand{\thefigure}{S\arabic{figure}}
\renewcommand{\thetable}{S\arabic{table}}

\section{Supplementary Materials}

\subsection{Cityscapes}
\label{app: cityscapes}

The encoder is a ResNet-50 model with dilated convolutions~\cite{yu2015multi}, pre-trained on ImageNet. We use a PSP module~\cite{he2015spatial} for the task-specific decoders. Every input image is rescaled to 512 x 256 pixels. We upsample the output of the PSP decoders back to the input resolution during training. The outputs are upsampled to 2048 x 1024 pixels during testing. The semantic segmentation task is learned with a weighted pixel-wise cross-entropy loss. We reuse the approach from~\cite{kendall2018multi} for the instance segmentation task, i.e. we consider the proxy task of regressing each pixel to the center of the instance it belongs to. The depth estimation task is learned using an L1 loss. The losses are normalized to avoid having the loss of one task overwhelm the others during training. The hyperparameters were optimized with a grid search procedure to ensure a fair comparison across all compared approaches.

\paragraph{Single-task models} We tested batches of size 4, 6 and 8, poly learning rate decay vs step learning rate decay with decay factor 10 and step size 30 epochs, and Adam (initial learning rates 2e-4, 1e-4, 5e-5, 1e-5) vs stochastic gradient descent with momentum 0.9 (initial learning rates 5e-2, 1e-2, 5e-3, 1e-3). This accounts for 48 hyperparameter settings in total. We repeated this procedure for every single task (semantic segmentation, instance segmentation and monocular depth estimation). 

\paragraph{Baseline multi-task network} We train with the same set of hyperparameters as before, i.e. 48 settings in total. We calculate the multi-task performance in accordance with~\cite{maninis2019attentive}. In particular, the multi-task performance of a model $m$ is measured as the average per-task performance increase/drop w.r.t. the single task models $b$:
\begin{equation}
    \label{eq: multi_task_performance}
    \Delta_{m} = \frac{1}{T} \sum_{i=1}^{T} \left(-1\right)^{l_i} \left( M_{m,i} - M_{b,i} \right) / M_{b,i},
\end{equation}
where $l_i = 1$ if a lower value means better for measure $M_i$ of task $i$, and $0$ otherwise.

\begin{figure}[H]
\begin{center}
\begin{minipage}[t]{.33\linewidth}
  \begin{center}
  \includegraphics[width=\textwidth]{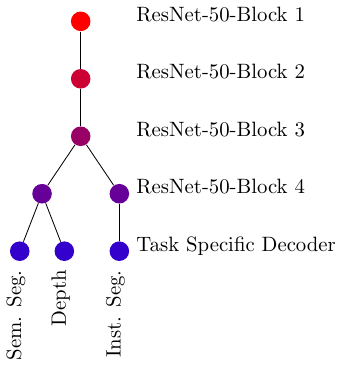}
  \end{center}
  \caption{Ours - 1}
  \label{fig:cityscapes1}
\end{minipage}%
\begin{minipage}[t]{.33\linewidth}
  \begin{center}
  \includegraphics[width=\textwidth]{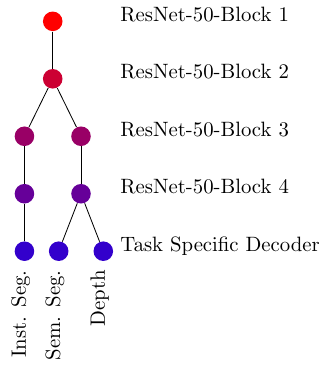}
  \end{center}
  \caption{Ours - 2}
  \label{fig: cityscapes2}
\end{minipage}
\begin{minipage}[t]{.33\linewidth}
  \begin{center}
  \includegraphics[width=\textwidth]{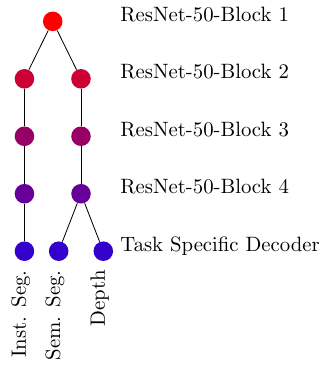}
  \end{center}
  \caption{Ours - 3}
  \label{fig: cityscapes3}
\end{minipage}
\end{center}
\caption{Branched multi-task networks on Cityscapes that were generated by our method.}
\label{fig: tree_cityscapes}
\end{figure}

\paragraph{Branched multi-task network} We reuse the hyperparameter setting with the best result for the baseline multi-task network. The branched multi-task architectures that were used for the quantitative evaluation on Cityscapes are shown in Fig.~\ref{fig: tree_cityscapes}.

\paragraph{Cross-stitch networks / NDDR-CNN} We insert a cross-stitch/NDDR unit after every ResNet block. We also tried to leave out the cross-stitch/NDDR unit after the final ResNet block, but this decreased performance. We tested two different initialization schemes for the weights in the cross-stitch/NDDR units, i.e. $\alpha = 0.8$, $\beta = 0.1$ and $\alpha = 0.9$, $\beta = 0.05$. The model weights were initialized from the set of the best single-task models above. We found that the Adam optimizer broke the initialization and refrained from using it. The best results were obtained with stochastic gradient descent with initial learning rate 1e-3 and momentum 0.9. As also done in~\cite{misra2016cross,gao2019nddr}, we set the weights of these units to have a learning rate that is 100 times higher than the base learning rate.

\paragraph{MTAN} We re-implemented the MTAN model~\cite{liu2019end} using a ResNet-50 backbone based on the code that was made publicly available by the authors. We obtained our best results when using an Adam optimizer. Other hyperparameters where set in accordance with our other experiments.  


\subsection{Taskonomy}
\label{app: taskonomy}

We reuse the setup from~\cite{zamir2018taskonomy}. All input images were rescaled to 256 x 256 pixels. We use a ResNet-50 encoder and replace the last stride 2 convolution by a stride 1 convolution. A 15-layer fully-convolutional decoder is used for the pixel-to-pixel prediction tasks. The decoder is composed of five convolutional layers followed by alternating convolutional and transposed convolutional layers. We use ReLU as non-linearity. Batch normalization is included in every layer except for the output layer. We use Kaiming He's initialization for both encoder and decoder. We use an L1 loss for the depth (D), edge detection (E) and keypoint detection (K) tasks. The scene categorization task is learned with a KL-divergence loss. We report performance on the scene categorization task by measuring the overlap in top-5 classes between the predictions and ground truth.

The multi-task models were optimized with task weights $w_s = 1, w_d = 1, w_k = 10, w_e = 10$ and $w_c = 1$. Notice that the heatmaps were linearly rescaled to lie between 0 and 1. During training we normalize the depth map by the standard deviation. 

\paragraph{Single-task models} We use an Adam optimizer with initial learning rate 1e-4. The learning rate is decayed by a factor of10 after 80000 iterations. We train the model for 120000 iterations. The batch size is set to 32. No additional data augmentation is applied. The weight decay term is set to 1e-4.

\paragraph{Baseline multi-task model} We use the same optimization procedure as for the single-task models. The multi-task performance is calculated using Eq.~\ref{eq: multi_task_performance}.

\begin{figure}
\begin{center}
\begin{minipage}[t]{.33\linewidth}
  \begin{center}
  \includegraphics[width=\textwidth]{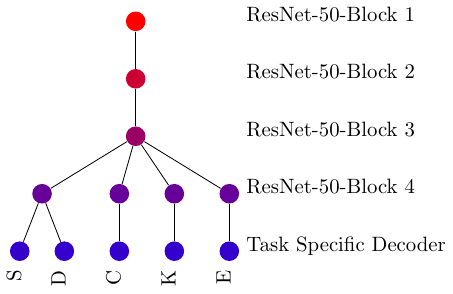}
  \end{center}  
  \caption{Ours - 1}
  \label{fig: taskonomy1}
\end{minipage}%
\begin{minipage}[t]{.33\linewidth}
  \begin{center}
  \includegraphics[width=\textwidth]{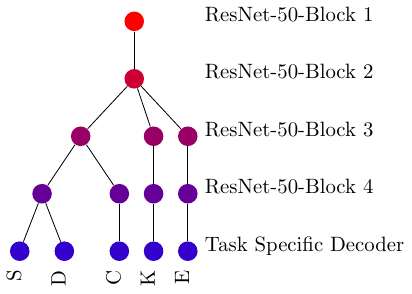}
  \end{center}  
  \caption{Ours - 2}
  \label{fig: taskonomy2}
\end{minipage}
\begin{minipage}[t]{.33\linewidth}
  \begin{center}
  \includegraphics[width=\textwidth]{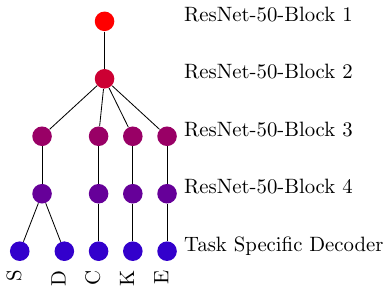}
  \end{center}  
  \caption{Ours - 3}
  \label{fig: taskonomy3}
\end{minipage}
\end{center}
\caption{Task groupings generated by our method on the Taskonomy dataset.}
\label{fig: our_tree_taskonomy}
\end{figure}

\begin{figure}[H]
\begin{center}
\begin{minipage}[t]{.33\linewidth}
  \begin{center}
  \includegraphics[width=\textwidth]{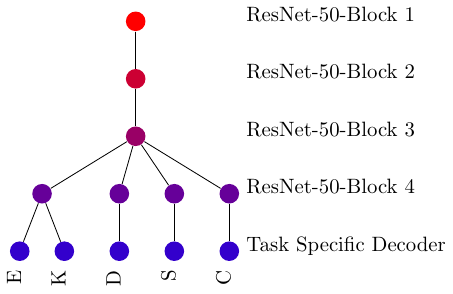}
  \end{center}
  \caption{FAFS - 1}
  \label{fig:FAtaskonomy1}
\end{minipage}%
\begin{minipage}[t]{.33\linewidth}
  \begin{center}
  \includegraphics[width=\textwidth]{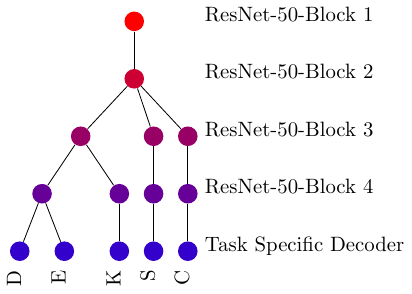}
  \end{center}
  \caption{FAFS - 2}
  \label{fig:FAtaskonomy2}
\end{minipage}
\begin{minipage}[t]{.33\linewidth}
  \begin{center}
  \includegraphics[width=\textwidth]{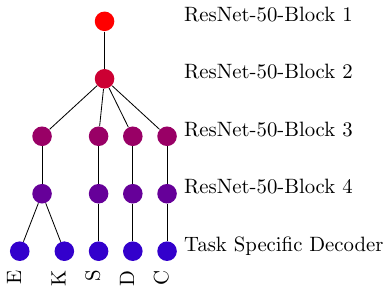}
  \end{center}
  \caption{FAFS - 3}
  \label{fig:FAtaskonomy3}
\end{minipage}
\end{center}
\caption{Task groupings generated on the Taskonomy dataset using the FAFS method from~\cite{lu2017fully}.}
\label{fig: fa_tree_taskonomy}
\end{figure}

\begin{figure}[H]
    \begin{center}
    \begin{minipage}[t]{\linewidth}
        \begin{center}
        \includegraphics[width=0.99\textwidth]{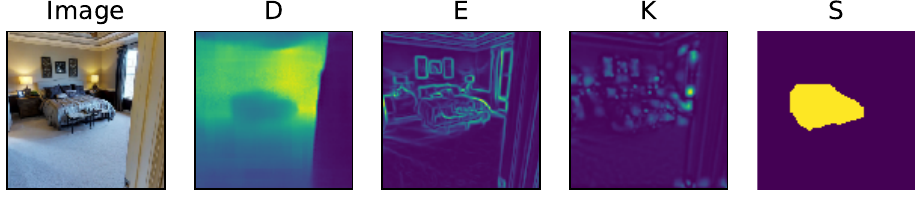}
        \end{center}
        \caption{Example - 1}
    \end{minipage}
    \begin{minipage}[t]{\linewidth}
        \begin{center}
        \includegraphics[width=0.99\textwidth]{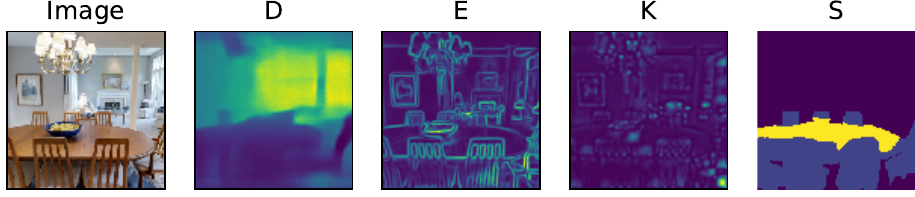}
        \end{center}
        \caption{Example - 2}
    \end{minipage}
        \begin{minipage}[t]{\linewidth}
        \begin{center}
        \includegraphics[width=0.99\textwidth]{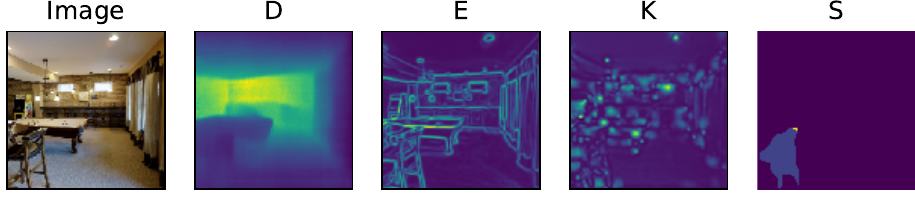}
        \end{center}
        \caption{Example - 3}
    \end{minipage}
    \end{center}
    \caption{Predictions made by our branched multi-task network on images from the Taskonomy test set.}
    \label{fig: taskonomy_qualitative}
\end{figure}

\paragraph{Branched multi-task models} We use the same optimization procedure as for the single-task models. The architectures that were generated by our method are shown in Fig.~\ref{fig: our_tree_taskonomy}. Fig.~\ref{fig: fa_tree_taskonomy} shows the architectures that are found when using the task grouping method from~\cite{lu2017fully}. We show some of the predictions made by our third branched multi-task network in Fig.~\ref{fig: taskonomy_qualitative} for the purpose of qualitative evaluation. 

\paragraph{Cross-stitch networks / NDDR-CNN} We reuse the hyperparameter settings that were found optimal on Cityscapes. Note that, these are in agreement with what the authors reported in their original papers. The weights of the cross-stitch/NDDR units were initialized with $\alpha=0.8$ and $\beta=0.05$.

\paragraph{MTAN} Similar to the other models, we reused the hyperparameter settings that were found optimal on Cityscapes. 

\subsection{CelebA}
\label{app: celeba}

We reuse the thin-$\omega$ model from~\cite{lu2017fully}. The CNN architecture is based on the VGG-16 model~\cite{simonyan2015very}. The number of convolutional features is set to the minimum between $\omega$ and the width of the corresponding layer in the VGG-16 model. The fully connected layers contain $2 \cdot \omega$ features. We train the branched multi-task network using stochastic gradient descent with momentum $0.9$ and initial learning rate $0.05$. We use batches of size $32$ and weight decay $0.0001$. The model is trained for $120000$ iterations and the learning rate divided by $10$ every $40000$ iterations. The loss function is a sigmoid cross-entropy loss with uniform weighing scheme. 

\begin{figure}[H]
\includegraphics[width=\linewidth]{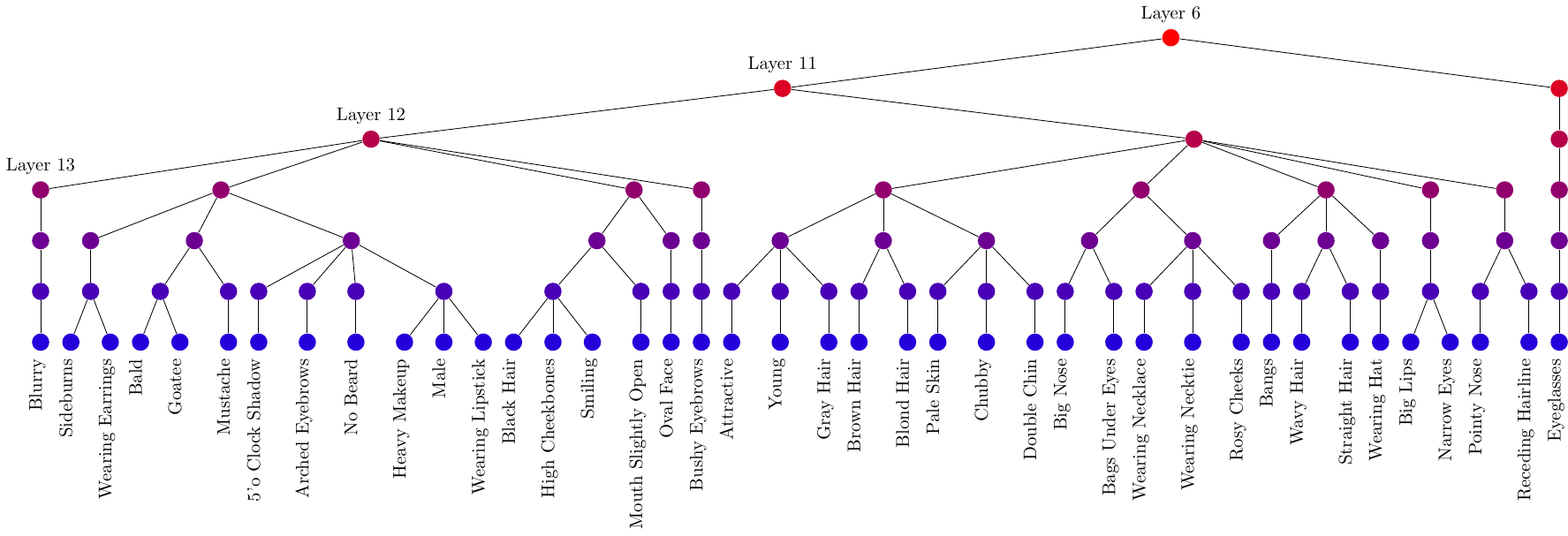}
\caption{Grouping of 40 person attribute classification tasks on CelebA in a thin VGG-16 architecture.}
\label{fig: tree_celeba}
\end{figure}

\subsection{Computational Analysis}

We provide an analysis to identify the computational costs related to the different steps when calculating the task affinity scores. We adopt the notation from the main paper. The following three steps can be identified:

\begin{itemize}
    \item Train $N$ single task networks. It is possible to use a subset of the available training data to reduce the training time. We verified that using a random subset of 500 train images on Cityscapes resulted in the same task groupings. 
    \item Compute the RDM matrix for all $N$ networks at $D$ pre-determined layers. This requires to compute the features for a set of $K$ images at the $D$ pre-determined layers in all $N$ networks. The $K$ images are usually taken as held-out images from the train set. We used $K =500$ in our experiments. In practice this means that computing the image features comes down to evaluating every model on $K$ images. The computed features are stored on disk afterwards. The RDM matrices are calculated from the stored features. This requires to calculate $N \times D \times K \times K$ correlations between two feature vectors (can be performed in parallel). We conclude that the computation time is negligible in comparison to training the single task networks. 
    \item Compute the RSA matrix at $D$ locations for all $N$ tasks. This requires to calculate $D \times N \times N$ correlations between the lower triangle part of the $K \times K$ RDM matrices. The computation time is negligible in comparison to training the single task networks. 
\end{itemize}

We conclude that the computational cost of our method boils down to training N single task networks plus some overhead. Notice that cross-stitch networks~\cite{misra2016cross} and NDDR-CNNs~\cite{gao2019nddr} also pre-train a set of single-task networks first, before combining them together using a soft parameter sharing mechanism. We conclude that our method only suffers from minor computational overhead compared to these methods. 

\bibliography{bibliography}
\end{document}